\documentclass[letterpaper, 10 pt, conference]{ieeeconf}  

\IEEEoverridecommandlockouts                              

\usepackage{xcolor}

\usepackage{graphicx}
\usepackage{subcaption}
\usepackage{float}
\usepackage{caption}
\usepackage{lscape}                                         

\usepackage[lined,ruled,linesnumbered]{algorithm2e}

\usepackage{booktabs}                   
\usepackage{multirow}

\usepackage{paralist}

\usepackage{bm}                          
\usepackage{epsfig}                      
\usepackage{graphicx}                  
\usepackage{times}
\usepackage{mathptmx}
\usepackage{mathtools}
\usepackage{amssymb,amsmath}   
\usepackage{scrextend}
\usepackage{tablefootnote}

\usepackage{units}

\usepackage{eucal}

\usepackage{comment}

\usepackage{url}  
\usepackage{xspace}
\usepackage{setspace}

\usepackage{layout}

\usepackage{amsmath,amsfonts,bm}

\def\eqref#1{equation~\ref{#1}}

\def\1{\bm{1}}

\DeclareMathAlphabet{\mathsfit}{\encodingdefault}{\sfdefault}{m}{sl}
\SetMathAlphabet{\mathsfit}{bold}{\encodingdefault}{\sfdefault}{bx}{n}

\renewcommand{\baselinestretch}{1}       

\newlength\paramargin
\newlength\figmargin
\newlength\secmargin
\newlength\figcapmargin

\long\def\ignorethis#1{}

\newcommand{\comments}{0}

\ifthenelse{\equal{\comments}{1}}
{
\newcommand {\yencheng}[1]{{\color{magenta}\textbf{Yen-Cheng: }#1}\normalfont}
\newcommand {\zk}[1]{{\color{blue}\textbf{Zsolt: }#1}\normalfont}
\newcommand {\jtian}[1]{{\color{orange}\textbf{JTian: }#1}\normalfont}
\newcommand {\cwkuo}[1]{{\color{blue}\textbf{Albert: }#1}\normalfont}
\newcommand {\cyma}[1]{{\color{cyan}{\textbf{Kevin: }#1}\normalfont}}
}
{
\newcommand {\zk}[1]{{\color{blue}{}}\normalfont}
\newcommand {\jtian}[1]{{\color{blue}{}}\normalfont}
\newcommand {\cwkuo}[1]{{\color{blue}{}}\normalfont}
\newcommand {\cyma}[1]{{\color{cyan}{{}}\normalfont}}
}

\newcommand{\final}{0}
\ifthenelse{\equal{\final}{1}}
{
\renewcommand{\yencheng}[1]{}
\renewcommand{\cwkuo}[1]{}
\renewcommand{\jtian}[1]{}
\renewcommand{\zsolt}[1]{}
\renewcommand{\cyma}[1]{}
}
{}

\newcommand{\DatasetName}{AirSim-CP}

\usepackage{biblatex} 
\usepackage{booktabs}
\usepackage{cleveref}

\title{\LARGE \bf Who2com: Collaborative Perception via Learnable Handshake Communication }

\author{
  Yen-Cheng Liu, Junjiao Tian, Chih-Yao Ma, Nathan Glaser, Chia-Wen Kuo and Zsolt Kira\\
Georgia Institute of Technology\\
\texttt{$\{$ycliu,jtian73,cyma,nglaser3,albert.cwkuo,zkira$\}$@gatech.edu}
}

\begin{document}
\maketitle


\begin{abstract}

In this paper, we propose the problem of \textit{collaborative perception}, where robots can combine their local observations with those of neighboring agents in a learnable way to improve accuracy on a perception task. Unlike existing work in robotics and multi-agent reinforcement learning, we formulate the problem as one where learned information must be shared across a set of agents \textit{in a bandwidth-sensitive manner} to optimize for scene understanding tasks such as semantic segmentation. Inspired by networking communication protocols, we propose a multi-stage handshake communication mechanism where the neural network can learn to compress relevant information needed for each stage. Specifically, a target agent with degraded sensor data sends a compressed request, the other agents respond with matching scores, and the target agent determines \textit{who to connect with} (\textit{i.e.,} receive information from). We additionally develop the \DatasetName~dataset and metrics based on the AirSim simulator where a group of aerial robots perceive diverse landscapes, such as roads, grasslands, buildings, \textit{etc}. We show that for the semantic segmentation task, our handshake communication method significantly improves accuracy by approximately 20\% over decentralized baselines, and is comparable to centralized ones using a quarter of the bandwidth. 
\end{abstract}


\section{Introduction}
A great deal of progress has been made in single-agent scene understanding using deep neural networks~\cite{Zhao_2017_CVPR,godard2017unsupervised,Chen_2019_CVPR}. 
However, as these methods become ubiquitous and larger numbers of robots are deployed in the world, it becomes beneficial for them to share knowledge via communication. For example, knowledge sharing across a fleet of self-driving cars could alleviate a number of challenges such as occlusion and sensor degradations or failures.

\begin{figure}[t!]
\centering
\includegraphics[width=8.5cm]{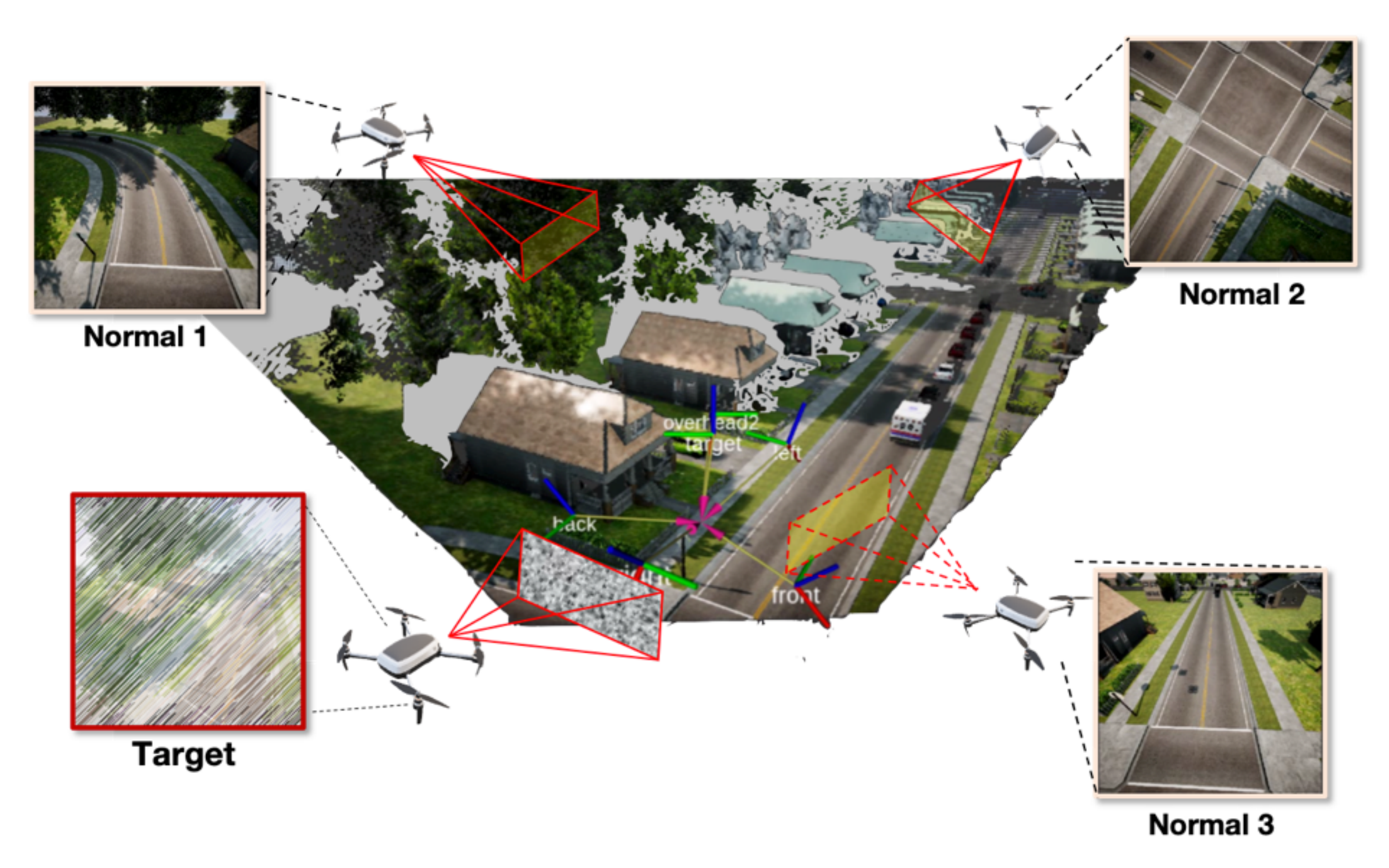}
\caption{
    \textbf{Illustration of collaborative perception.} Our collaborative perception task is to improve the perception ability of a degraded agent using information from  other agents in a bandwidth-limited way. 
}
\label{fig:overview}
\vspace{-4mm}
\end{figure}

In this paper, we propose the problem of \textit{collaborative perception}, where robots can combine their local observations with those of neighboring agents to improve accuracy in a perception task, such as semantic segmentation~\cite{long2015fully,chen2017rethinking}. 
This can result in significant improvements, for example when the receiving agent's sensors are occluded or degraded (see Figure~\ref{fig:overview}). We therefore formulate a problem where a degraded agent can communicate with other agents to improve its perceptual abilities. 
Unlike past methods that focus on multi-robot localization and mapping~\cite{cunningham2010ddf, saeedi2016multiple}, we develop agents that \textit{learn what to communicate} in a manner that is amenable to end-to-end deep learning, which dominates scene understanding.
In addition, different from multi-agent reinforcement learning~\cite{sukhbaatar2016learning,battaglia2016interaction,foerster2016dial,peng2017multiagent,jiang2018learning,pesce2019improving,foerster2018counterfactual}, we seek to do so under bandwidth constraints. We therefore propose to learn \textit{who to communicate with} in order to reduce bandwidth requirements while improving accuracy. Such a method is effective for standard observation problems which have the sub-modularity property, \textit{i.e.,} that adding more observation agents achieves diminishing returns~\cite{krause2007near}. Similar advantages have also been shown using rate distortion theory~\cite{dobbe2017fully}. 

In order to investigate the inherent trade-off between accuracy and bandwidth, especially in a manner that scales in a bounded way with respect to the number of agents, we propose a three-stage communication mechanism inspired by three-way handshaking in the regime of communication networking~\cite{kurose2005computer}. 
\begin{figure*}[t!]
\vspace{2mm}

\centering
\includegraphics[width=1\linewidth]{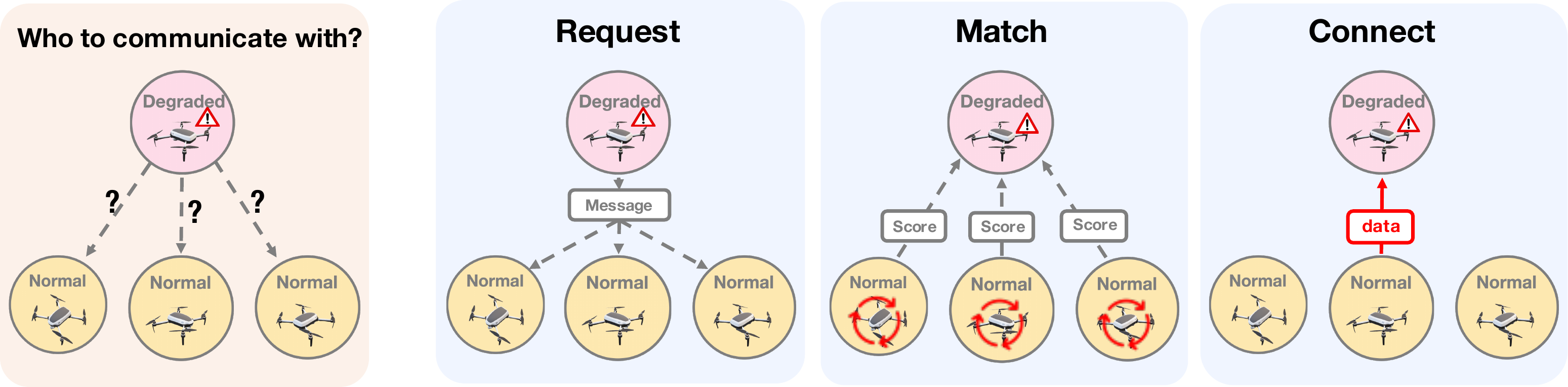}
\caption{
    \textbf{The concept of three-stage handshake communication.} Our model consists of three major steps to establish the connection with the selected agent. 
    \textbf{Request}: the degraded agent first broadcasts the compressed request message to other agents.
    \textbf{Match}: the normal agents compute matching scores based on the individual observations and the received request message.
    \textbf{Connect}: after returning all scores back to the target degraded agent, it connects to one of the normal agents with the highest matching score and  obtains  high-bandwidth data (\textit{i.e.,} feature maps for the semantic segmentation task) from it. 
}
\label{fig:concept}
\vspace{-4mm}
\end{figure*}
The three steps of our method are: 1) \textbf{request}: the degraded agent broadcasts a compressed request conditioned on its visual observation, 2) \textbf{match}: the other agents compute a learned matching score between their own visual observations and the received request, and 3) \textbf{connect}: the degraded agent selects one of the agents to communicate with and further improve its own prediction accuracy in downstream perception tasks.
The entire mechanism is trained in an end-to-end manner, using only supervision for the down-stream task (\textit{e.g.,} semantic segmentation). We show that this communication mechanism effectively decouples the request, score, and actual transmitted value to allow different (\textit{i.e.,} asymmetric) sizes during this communication. Our experiments demonstrate that this results in significant bandwidth savings when compared to centralized baselines and accuracy gains over uniform compression across agents.

In order to investigate the properties of the resulting method, we develop the AirSim-CP dataset and metrics using the AirSim simulator~\cite{airsim2017fsr}, where a group of aerial robots fly over a map with diverse landscapes, such as roads, grasslands, buildings, lakes, \textit{etc}.
We vary a number of elements across the scenarios, including their trajectories, partial or complete overlap between the fields of view of the target and other agents, and with noisy or accurate pose information available for cross-view geometric warping. 
We quantitatively show that our proposed method is able to significantly improve accuracy by approximately 20\% over decentralized baselines, and is comparable to centralized ones using a quarter of the bandwidth.

We highlight the contributions of this paper as follows:

\begin{itemize}

\item We propose a new problem in the intersection of multi-agent systems, perception, communication, and learning. Compared with prior works on learning to communicate, we are the first to tackle the problem of learning to communicate with bandwidth constraints to the best of our knowledge. 

\item  Different from other multi-agent systems~\cite{hoshen2017vain,jiang2018learning}, our collected dataset, AirSim-CP, provides high-resolution and photo-realistic images for better evaluating multi-agent perception tasks with communication.

\item We propose an end-to-end communication framework trained without supervision indicating the ground-truth agent for communication, and shows superior accuracy to decentralized baselines and comparably to strong centralized ones with a fraction of the bandwidth.

\end{itemize}


\section{Related Work}
\label{sec:related_work}

Communication in multi-agent environments is a foundation of both collaborative perception and decision-making. 
This topic has been studied extensively in the regime of Multi-Agent Reinforcement Learning (MARL)~\cite{littman1994markov,schmidhuber1996general,panait2005cooperative}. Early attempts utilized a pre-defined communication protocol to propagate the information across agents~\cite{tan1993multi,zhang2013coordinating,melo2011querypomdp}, while dynamics of environments and a varied number of agents gave rise to the development of learnable mulit-agent communication~\cite{sukhbaatar2016learning,foerster2016learning,battaglia2016interaction,hoshen2017vain,foerster2016dial,peng2017multiagent,jiang2018learning,singh2018learning,kim2018learning,das2019tarmac,pesce2019improving,foerster2018counterfactual,jain2019two,kim2019message}. 
Existing works on MARL demonstrated the effectiveness of communication for various tasks, applying the models on simplistic 2D grid environments where the  observation of each agent is low-dimensional. 
As noted in Jain~\textit{et al.}~\cite{jain2019two}, studying collaboration in simplistic environments does not permit study of the interplay between perception and communication. 
Thus, in this work, we examine our framework in a more complicated and photorealistic environment. 

\begin{figure*}[t]
    \begin{center}
    \centerline{\includegraphics[width=0.6\linewidth]{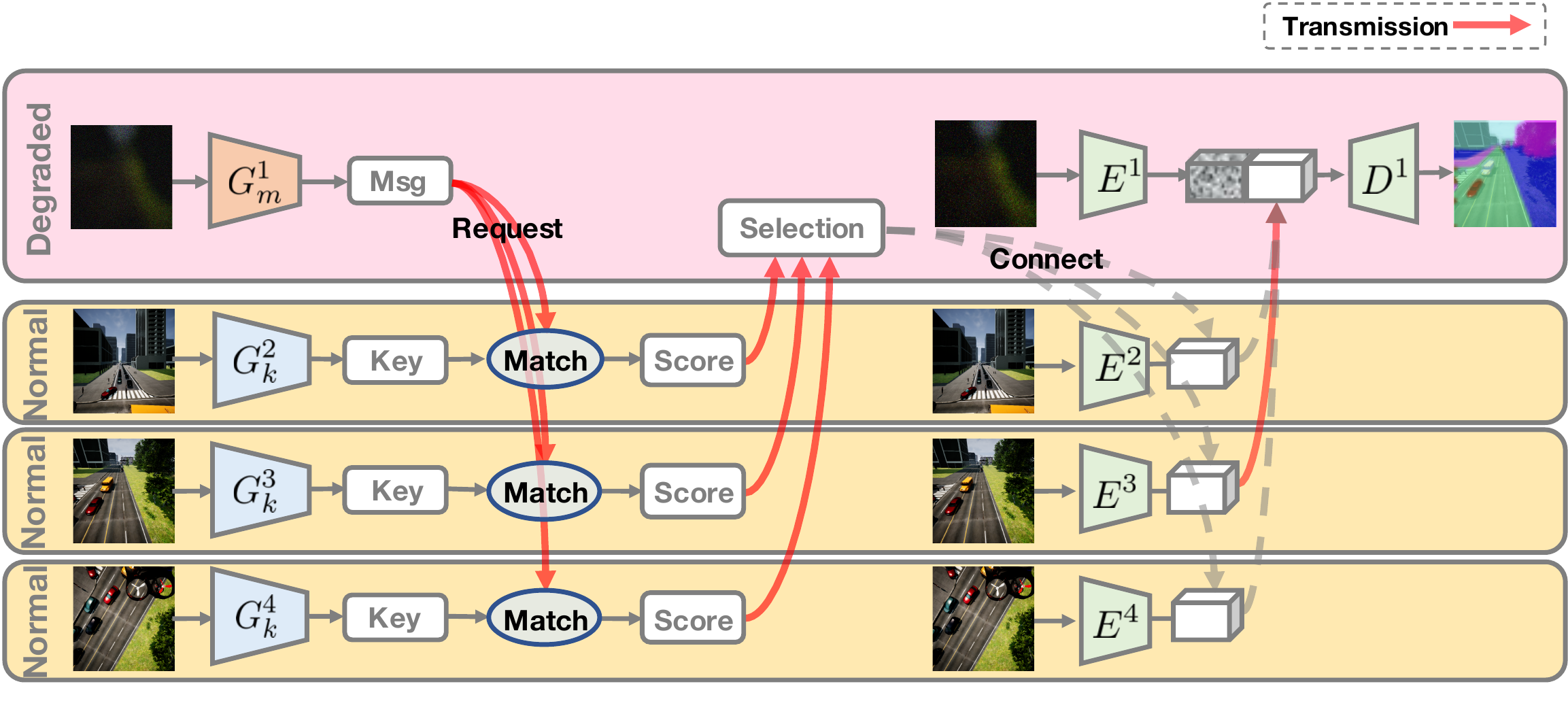}}
    \caption{
        \textbf{Our proposed model and communication steps.}
        The degraded agent first computes a low-dimensional vector as the request message and broadcasts it to other normal agents, and each normal agent then generates a matching score conditioned on the request message. 
        Finally, the degraded agent accesses the requested information from one of the agents during inference (and from all agents during training).
        The proposed method is trained end-to-end and significantly improves accuracy over decentralized baselines while minimizing bandwidth usage over centralized ones.
    }
    \label{fig:arch}
    \end{center}
\vspace{-6mm}
\end{figure*}

Among existing works on learning to communicate, the most relevant framework to our work is TarMac~\cite{das2019tarmac}, where the targeted communication is defined as the transmitted message determined by both the sender and receiver agents. 
However, during the communication, both message and data are broadcast to all of the other agents, hence not taking bandwidth usage into account. 
On the other hand, other recent work proposed to construct the communication group based on 
either pre-defined rules~\cite{jiang2018learning,jiang2018graph} or a unified communication network~\cite{singh2018learning,sukhbaatar2016learning,hoshen2017vain,peng2017multiagent,singh2018learning,das2019tarmac}. 
In this way, the bandwidth usage during communication increases as the number of agents scales up. In contrast, our framework aims to minimize the bandwidth consumption yet maintain performance for the perception task by selecting which agent(s) to communicate with.

\section{Proposed Method}

\subsection{Problem Definition and Motivation}
\cyma{Need an opening here for the following paragraph. Perhaps the opening can be even before Sec. 3.1. Right now, it's not clear what is our environment.}
In our proposed \textit{collaborative perception} task, an environment consists of $N$ agents with their own observations $\bm{X} = \{\bm{x}_i\}_{i=1,...,N}$, while the observation of the target agent $\tilde{\bm{x}}_j, j\in\{1,...,N\}$ is degraded. The goal of the target agent is to integrate information received from other agents to derive the prediction for its own local observation. 
Realistically, communication mechanisms often have bandwidth limitations, preventing the transmission of a large quantity of information during communication.\cyma{missing citation.}
Thus, our goal is to derive a distributed and information-fusing framework which is able to (1) maximize the prediction accuracy of the downstream perception tasks for the target agent and (2) minimize the bandwidth used during transmission, as illustrated in Figure~\ref{fig:overview}. 
We propose a communication framework that can be generalized to many perception tasks, and in this paper we consider semantic segmentation as one instantiation to evaluate the framework.

\subsection{Communication via Three-Stage Handshake}
Rather than broadcasting all information within the entire network in a brute-force manner, an efficient way of minimizing the bandwidth usage while  maintaining performance is to select which agent to communicates with. Toward this end, inspired by related work in communication networks~\cite{kurose2005computer}, we introduce a three-stage handshake communication mechanism 
shown in Figure~\ref{fig:concept}. Such a selection is motivated by works on rate distortion theory~\cite{dobbe2017fully} and sub-modularity of multiple observations~\cite{krause2007near}. Our empirical experimental results also show that this approach yields a better trade-off between bandwidth usage and performance, compared to uniform compression of all information from the agents.\cyma{add co-reference to the experiment section.}

Our communication mechanism consists of three major steps: \textbf{request}, \textbf{match}, and \textbf{connect}. Specifically, the degraded agent first broadcasts its request message, $\bm{\mu}_j \in\mathbb{R}^m$, to the neighboring normal agents, and the normal agents compute a matching score, $s_{ji}$, between their keys, $\bm{\kappa}_i \in\mathbb{R}^k$, and the request message. 
Once the normal agents return their matching scores back to the degraded agent, the degraded agent further selects the best $n$ agents to connect with (\textit{i.e.,} receive information from) according to these matching scores\footnote{Note that in this paper we select the top $n=1$, but the method can be generalized to top-$n$ selection.}. 
The complete communication steps are illustrated in Figure~\ref{fig:arch}.
In each step, information can be compressed by each agent $i$ through a key generator $G^i_k$, a message generator  $G^i_m$, an image encoder $E^i$, and an task decoder $D^i$. \cyma{Image decoder is specific for semantic segmentation. Let us see if we can sell it in a way that is more generalizable and immediately say that for the sake of using semantic segmentation, we will need an image decoder.}\zk{changed to task decoder?}
Note that this approach effectively decouples the various stages, allowing for different compression rates for the message, keys, and values. This significantly benefits the trade-off between bandwidth and accuracy. We now detail the steps: 

\textit{\textbf{Request}} - 
The degraded agent $j$ first compresses its observation $\tilde{\bm{x}}_j$ to a low-dimensional message $\bm{\mu}_j$:
\begin{equation}
     \bm{\mu}_j = G^j_m(\tilde{\bm{x}}_j; \theta_m)\in\mathbb{R}^m,
\end{equation}
where $G^j_m$ is a message generator parameterized by $\theta_m$. The propagated message $\bm{\mu}_j$ compresses important information from the local observation of the degraded agent $j$.

\textit{\textbf{Match}} - 
In the match step, each of other agents derives the matching score $s_{ji}$ between the received request message $\bm{\mu}_j$ from the degraded agent and the key $\bm{\kappa}_i$ generated from its own observations,
\begin{equation}
     s_{ji} = \Phi(\bm{\mu}_j, \bm{\kappa}_i), \quad \bm{\kappa}_i = G^i_k(\bm{x}_i;\theta_k)\in\mathbb{R}^k,
\end{equation}
where $\Phi(\cdot, \cdot)$ represents the matching function of two vectors and $G^i_k$ denotes a key generator parameterized by $\theta_k$. 
We use the \textit{general} attention mechanism~\cite{luong2015effective} as our matching function:
\begin{align}\label{eq:attention}
    \text{General:}\ \Phi &= \bm{\mu}_j^T \bm{W}_a \bm{\kappa}_i,
\end{align}
where $\bm{W}_a$ is a learnable parameter. 
We also compare it with two other attention mechanisms: $\text{Scale Dot-Product}\ \Phi = \bm{\mu}_j^T\bm{\kappa}_i/\sqrt{d_n}$~\cite{vaswani2017attention} and $\text{Additive}\ \Phi = \bm{W}_a^T \text{tanh}(\bm{W}_k\kappa_i + \bm{W}_m \bm{\mu}_j)$~\cite{bahdanau2014neural},
where $\bm{W}_k, \bm{W}_m$ denote parameters to be learned and $d_n$ denotes the dimension of the message and keys. Note that only the general and additive functions allow for different key and message sizes. The scale dot-product function requires identical message-key size.
Empirically, we find that \textit{general} attention achieve the best performance in our experiments and hence use it unless otherwise specified. Note that we assume equal cost links between agents, though our models can further support per-link costs (unlike the baseline methods). We leave this for future work.

\textit{\textbf{Connect}} - 
In the connect step, the selected $\hat{i}$-th agent transmits the requested information (\textit{e.g.,} a feature map for semantic segmentation) $\bm{f}_{\hat{i}}$ to the degraded agent. 
With the integrated information, the target degraded agent makes a final prediction  $\tilde{y}_j$ as follows:
\begin{equation}\label{eq:inference}
\bm{\tilde{y}}_j=D^j([\tilde{\bm{f}}_j; \bm{f}_{\hat{i}} ];\theta_d),
\end{equation}
where $\bm{f}_{\hat{i}} = E^i(\bm{x}_{\hat{i}};\theta_e)\in\mathbb{R}^{d_f \times d_f \times d_c}$ is the feature map from the local observation of normal agent $\hat{i}$, $d_f, d_c$ are the spatial dimension and number of channels of the feature maps respectively, $\tilde{\bm{f}}_j = E^j(\bm{\tilde{x}}_j;\theta_e)$ is the feature map from the noisy observation of the degraded agent, and $[\cdot\; ; \; \cdot]$ is the concatenation operator along the channel dimension. $\theta_e$ and $\theta_d$ are the parameters in the task encoder and decoder.



\subsection{Learning to Communicate with Weak Supervision}
\label{sec: centralized-training}
\textbf{Centralized training with decentralized execution.}
Our learning strategy is inspired by the concept of centralized training with decentralized execution~\cite{lowe2017multi}. During training our target agent can access the observations of all other agents. 
On the other hand, during inference the target agent is required to perform in a bandwidth-limited manner by only accessing information from the selected agent(s). 

Specifically, during training, the task decoder uses the sum of observations from all normal agents weighted by their corresponding matching scores and further computes the final prediction akin to Eq.~\ref{eq:inference} during inference:
\begin{align}\label{eq:softmax}
        \bm{\tilde{y}}_j=D^j([\tilde{\bm{f}}_j; \bm{f}_{sum}];\theta_d), \quad
        \bm{f}_{sum} = \sum_{i=1}^N \alpha_{j,i}\bm{f}_i, \quad
\end{align}
where $\alpha_{j,i}$ is $i$th element of $ \alpha_j = \rho( \left[ s_{j1};...;s_{jN}\right] ) \in\mathbb{R}^{N}$ and $\rho$ is a softmax operation.
The most straightforward decentralized execution method is simply adopting \textit{argmax} selection, \textit{i.e.,} 
connecting only to the agent with the highest computed matching score:
\begin{equation}\label{eq:argmax}
     \hat{i} = \underset{i}{\mathrm{argmax}} \ s_{ji}.
\end{equation}
However, \textit{argmax} selection is non-differential during training. 
We address this issue by simply applying the \textit{softmax} operator in the training stage and \textit{argmax} operator in the inference stage. 
Empirically, we find that this simple method achieves similar results compared to other more complex schemes, \textit{e.g.,} \textit{sparsemax}~\cite{martins2016softmax}. \zk{Add this comparison if room?} Note also that this can be generalized to top-\textit{n} selection as well. 

\textbf{Training objective.} Learning our model does not require supervision of ground truth labels indicating the best agent to communicate with.
Therefore, the only supervision for our model comes from the ground truth annotation at the target view (\textit{e.g.,} segmentation masks). The objective function of our model, which is trained end-to-end, can thus be defined as $\mathcal{L} = \mathcal{H}(\bm{y}_j, \bm{\tilde{y}}_j),$
where $\mathcal{H}$ is the cross-entropy loss, and $\bm{y}_j$ denotes the ground truth labels of the target agent's view.

	

\begin{figure}[t]
    \vspace{1mm}
    \begin{center}
    \centerline{\includegraphics[width=\linewidth]{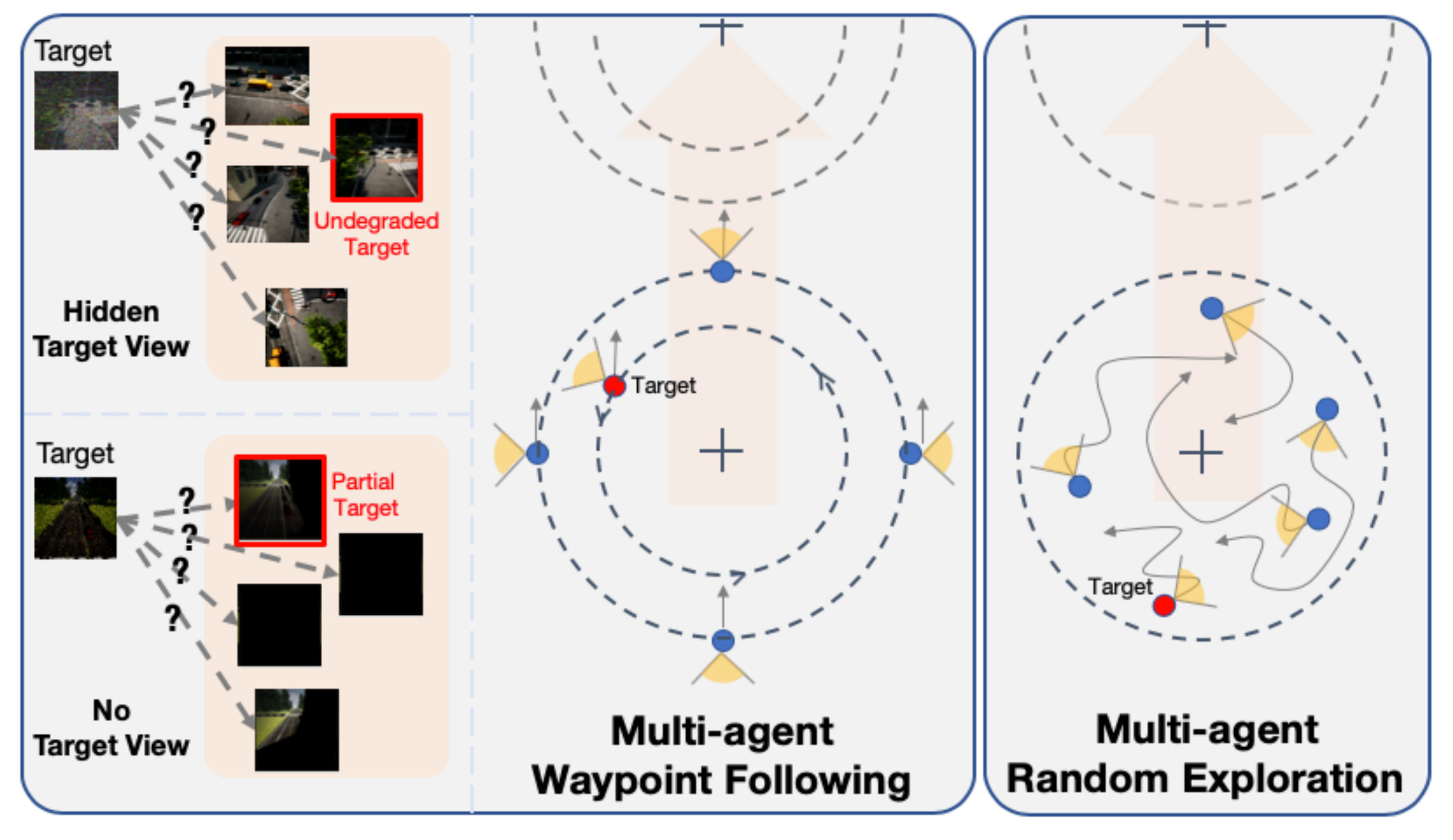}}
    \caption{
        \textbf{Illustration of our experimental cases.} 
    }
        \label{fig:case_des}

    \end{center}
    \vspace{-6mm}
\end{figure}

\section{Experimental Results}
\label{sec:result}

\subsection{AirSim-CP Dataset}
\textbf{Dataset.} Our \DatasetName~dataset is built upond the AirSim simulator~\cite{airsim2017fsr}, where a group of \textbf{five} drones fly over a map with diverse landscapes, such as roads, grasslands, buildings, lakes, \textit{etc}. 
Currently, in our \DatasetName~dataset, we use  semantic segmentation as the downstream task to benchmark methods for the collaborative perception problem. 
For each drone, RGB images, depth images, and poses are recorded. We also provide the semantic segmentation mask of one of agents. 

\subsection{Proposed Experimental Settings.}
\label{sec:setting}
As illustrated in Figure~\ref{fig:case_des}, in order to obtain perceptual data under realistic trajectories, the drones perform pre-specified tasks of waypoint following and multi-agent random exploration. 
We then consider \textbf{four} experimental settings: (1) hidden target view (waypoint following trajectories), (2-3) accurate or inaccurate pose (waypoint following trajectories), and (4) accurate pose (random exploration trajectories). We collect approximately 10-20k images per setting, with a roughly 60\%/20\%/20\% train/val/test split.
In all cases, there is a degraded target agent. 
We perturb the target agent's view by applying a Gaussian blur filter with random size (from $1$ to $100$) and Gaussian noise.
When specified, depth and pose information is used to warp the normal agents' views to the target view. In terms of the optimization, we use ResNet18~\cite{he2016deep} as our feature backbone and train it for $200k$ iterations with Adam optimizer with learning rate $10^{-5}$.
                                
\textbf{Hidden Target View with Multi-agent Waypoint Following.} 
The agents are asked to navigate between waypoints but when performing the semantic segmentation task, rather than the \textit{normal} agents being all of the neighboring agents, we replace one of the normal agent with an \textit{un-degraded} version of the target view. 
This task tests whether the baseline and the proposed method can help the target agent find the \textit{hidden} ground-truth target view in one of the neighboring agents. 
This experiment thus can be regarded as a sanity check on communication. 
Note that we do not use any 3D information to warp normal views to the target view. 
The motivation for this case is to remove the confound of geometric warping and image misalignments (which is important for the semantic segmentation task~\cite{ronneberger2015u})
from the study of the collaborative perception task.
Our focus is to make sure that communication can be effective and accurate in a bandwidth-limited manner.
Hence, we can directly assess only communication effectiveness from this case.

\textbf{Accurate Pose with Multi-agent Waypoint Following.}  
Similar to the previous setting, the five drones are performing waypoint following  jointly. 
Differently, the field-of-view (FOV) of the target agent only partially overlaps with some subset of the normal agents. 
This case is designed to test whether the proposed method can select the normal agent with partially overlapping FOVs, in order to aid the downstream perception task. 
In order to maintain the image alignment for a better segmentation prediction, we use 3D information from the depth maps of each normal agent's view and an accurate relative pose transformation to the target view to warp the pixels of the normal agent's observations to the target's view. Note that the depth maps do not have to be transmitted (warping is done locally on each agent) but we include the transmission of the target's pose to the other agents in the bandwidth used, although it is minimal.

\textbf{Inaccurate Pose with Multi-agent Waypoint Following.} To further make our experimental setting more realistic, we add noise in the  agents' positions. This results in warped images that are not well-aligned with the target view.

\textbf{Accurate Pose with Multi-agent Random Exploration.} We also investigate collaborative perception during multi-agent random exploration, where agents approach a target location, disperse, and wander. As the agents explore the environment individually, the relative positions and overlapping fields-of-view of agents change frequently.

\subsection{Baselines}
We consider the following methods for comparison:
\begin{itemize}
    \item \textit{Single normal (upper bound)} and \textit{Single degraded (lower bound) }: the models are trained with single non-degraded and degraded images respectively for the target agent.
    \item \textit{CatAll (centralized)}: the model uses the concatenation of all features from both degraded and normal agents as input for semantic segmentation.
    \item \textit{Attention (centralized)}:     the \textit{Attention} mechanism weights and sums up  feature maps instead of the concatenation of the \textit{CatAll} method. 
    \item \textit{Compression (centralized)}: the compression model applies two additional convolutional layers and performs uniform compression of all observations at rate $25\%$, with concatenation used for combining them. Note that we can certainly replace our image encoder with more sophisticated compression encoders to further improve the compression rate~\cite{cheng2019learning}.
    
    \item \textit{Random selection (distributed)}: 
    instead of learning to select which agent to communicate with, here the feature map from a random normal agent is selected.
    \item \textit{Ours (distributed)}: we denote our proposed method as ours with message (\textit{ours w/ msg}), and another variant where the message request $\bm{\mu}_j$ is set to a constant vector with ones to check whether the message request is essential. It is worth noting that we \textit{do not} use any label indicating the best agent during the training. 
\end{itemize}
Both \textit{CatAll} and \textit{Attention} require all feature maps from normal agents to be sent to the degraded agent. The bandwidth of the centralized baselines scales linearly with the number of agents in the system, while \textit{Random selection} and \textit{ours} only requires a single image feature map to be transmitted.

\subsection{Evaluation metrics.}
In order to evaluate and analyze the effectiveness of models, we use 1) overall accuracy to measure the performance of semantic segmentation~\cite{long2015fully} and 2) Kbytes per frame (kbpf) to measure the bandwidth usage (BW) to examine the ability of communication and selection. 
In addition, to better benchmark the performances on the collaborative perception task under limited bandwidth, we introduce the \textbf{Bandwidth-Improvement Score} (BIS) defined as:
\begin{equation}
    BIS = \dfrac{ \delta -\bar{\delta}}{(\hat{\delta}-\bar{\delta})\omega},
\end{equation}
where $\delta$ is the overall accuracy of the examined method, $\bar{\delta}$ is the overall accuracy of the single degraded model (\textit{i.e.,} lower bound on overall accuracy), $\hat{\delta}$ is the overall accuracy of the single normal model (\textit{i.e.,} upper bound on overall accuracy), and $\omega$ is the bandwidth usage (in Mbytes per frame) of the examined method. The BIS score is defined as a ratio of relative improvement in overall accuracy over bandwidth usage. Smaller bandwidth usage and larger improvements in overall accuracy lead to higher scores. 

\begin{table*}
\vspace{2mm}
\caption{\textbf{Experimental results on Multi-agent Waypoint Following and Random Exploration.} }
\label{tab:res_nav}
\centering
\resizebox{\linewidth}{!}{%
\begin{tabular}{cccccccccccccccccccc}
\toprule
& & &\multicolumn{2}{c}{Waypoint Following}& &\multicolumn{2}{c}{Waypoint Following}& & \multicolumn{2}{c}{Waypoint Following } & & \multicolumn{2}{c}{Random Exploration }\\

& & &\multicolumn{2}{c}{(Hidden Target View)}& &\multicolumn{2}{c}{(Accurate Pose)}& & \multicolumn{2}{c}{(Inaccurate Pose)} & & \multicolumn{2}{c}{(Accurate Pose)}  \\
\cmidrule(lr){4-5} \cmidrule(lr){7-8}\cmidrule(lr){10-11} \cmidrule(lr){13-14}

& & BW (kpbf) &Overall Acc & BIS& &Overall Acc & BIS& &Overall Acc & BIS& &Overall Acc & BIS \\
\midrule
Upper bound & Single Normal& -& 88.14   & -  & & 88.13  & -   & & 89.7 &-& & 89.16 &-\\
\midrule
\multirow{3}{*}{Centralized }  & CatAll & 4096 & 72.58 & 0.049& &80.05  & 0.17& & 80.33  & 0.172& & 78.74  & 0.133\\
& Attention  & 4096.03 & 69.08  &  0.004 & & 84.38  & 0.212 & & 82.79  & 0.174& & 81.07  & 0.159\\
& Compression &1024 & 70.44 &  0.085 & &76.2  & 0.527 & & 76.93  & 0.441& & 73.05  & 0.277\\
\midrule
\multirow{3}{*}{ Distributed }
& Random Selection & 1024 & 69.16  & 0.019 & & 64.94  & 0.082& & 67.49  & 0.028& & 69.58  & 0.123\\
& Ours w/o msg  & 1024.03 & 65.31  & -0.179 & & 78.10  & 0.602& &79.34  & 0.546& & 80.69  & 0.621\\
& Ours w/ msg & 1028.03 &84.57  &   \textbf{0.812} & & 80.42  &  \textbf{0.691} & &79.44  & \textbf{0.549}& & 80.97  & \textbf{0.631}\\
\midrule
Lower bound & Single Degraded & - & 68.79  &  -  & & 62.88 & -  & &66.84 & - & &66.85 & -  \\
\bottomrule
\end{tabular}}
\end{table*}

 \begin{figure*}[ht]
   \begin{picture}(0,110)
     \put(15,10){\includegraphics[width=8cm]{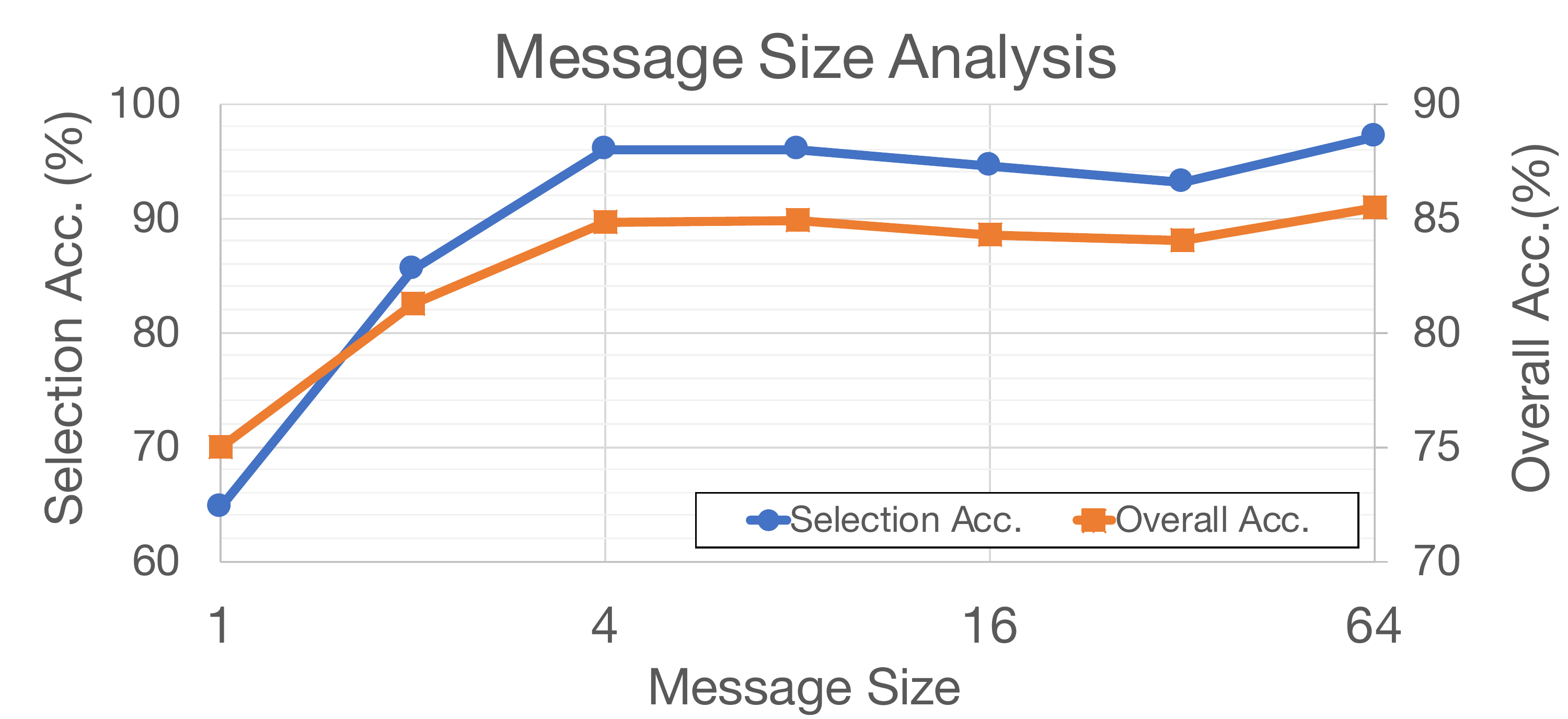}}
     \put(275,10){\includegraphics[width=8cm]{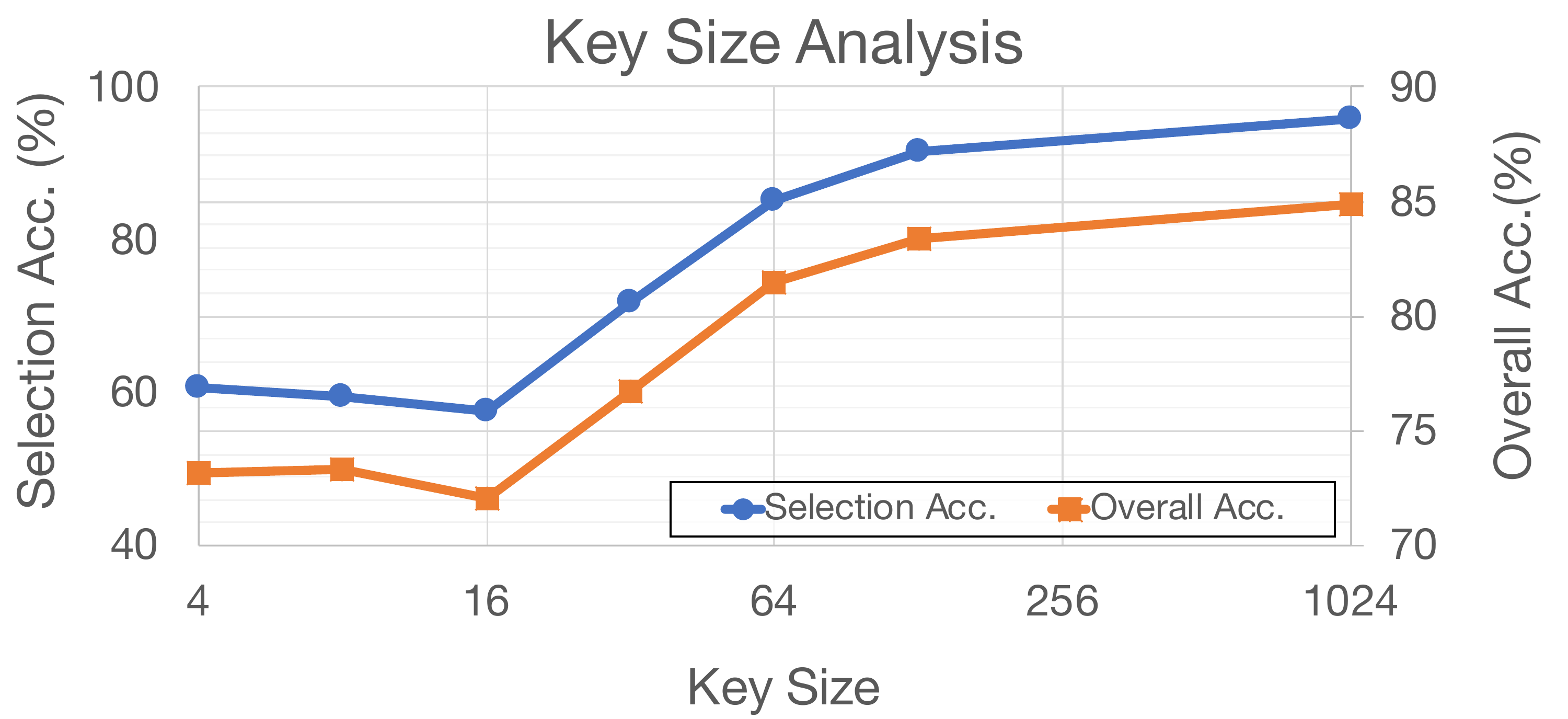}}
     \put(127,0){(a)}
     \put(383,0){(b)}

   \end{picture}
\caption{\textbf{Ablation study on varying (a) message and (b) key size.} We vary the size of key from $4$ to $1024$ and vary the size of message from $1$ to $64$ on Hidden Target View. Note that we use the key with size $8$ on the varying message analysis, while the message with size $1024$ is used for key size analysis. Without the message, the model performs the selection accuracy of $25.52\%$ and the overall accuracy of $61.4\%$, and we use key size of $1024$ for (a) and message size of $8$ for (b).}
\label{fig:message_key}
\vspace{-7mm}
\end{figure*}

\section{Results and Ablation Studies}
\label{sec:results-cases}

In this section, we compare our proposed method with baselines on the four cases (Sec.~\ref{sec:setting}) as shown in Table~\ref{tab:res_nav}.

\textbf{ Hidden Target View with Multi-agent Waypoint Following.}
As mentioned, geometric warping is not applied for this case. This results in better evaluation of communication, since the warping noise for the target view is removed and non-warped normal images make the selection more difficult.

Several observations can be made from this case.
First, as the centralized baselines are able to access all observations from different views, it should be the upper bounds of all distributed methods including our proposed models. However, we find that our model improves overall accuracy by a relative \textit{$16.51\%$} with respect to CatAll with only one quarter bandwidth usage. This shows that simply concatenating all of the information cannot guarantee that the network will combine it meaningfully and bandwidth is likely to be wasted. Second, in order to predict pixel-wise outputs and accurately predict fine-grained classes, scene understanding tasks require high-dimensional feature maps during inference. Using the overly compressed feature map may degrade the overall accuracy, hence our model with the message can improve overall accuracy a relative $20.06\%$ with respect to the \textit{Compression} model. Lastly, we also observe that our model with the message can improve mIoU by a relative $29.49\%$ compared to without the message. This demonstrates the necessity of the message in the communication.

\textbf{Accurate and Inaccurate Pose with Multi-agent Waypoint Following.}
One challenge for these cases is that the field of view between target and normal agents are partially overlapping. With the forward geometric warping on the visual observations of the normal agents, we observe that only one or two agents contain(s) partial information of the target view. Thus the performances of distributed models drop compared to the previous task. Our models are still able to achieve similar results compared with centralized methods, using only a quarter of the bandwidth, and still achieve the highest BIS among these methods. 
It is worth noting that our models with and without the message perform similarly because the model can rely on cues from warping (\textit{e.g.,} amount of overlap) from the normal agent images, hence reducing the need for conditioning on the message. Note also that although the BIS degrades somewhat due to inaccurate pose, it is still able to significantly beat the baselines.

\textbf{Accurate Pose with Multi-agent Random Exploration.}
Although overlapping FOVs and motions of agents change frequently in this case, we observe that our models still perform favorably against other baseline methods. This shows our models' robustness in different environments and tasks. 

\textbf{Message and key sizes.} 
To better examine the accuracy of agent selection, we manually label the "best" agent of the testing sets of Hidden Target View 
and further measure the selection accuracy and overall accuracy of various models. 

When using the \textit{general} attention, the size of message and key can be set to different values. We first analyze the effect of message size by varying it from $1$ to $64$ as shown in Figure~\ref{fig:message_key}a.
We observe that a larger message size results in increased selection accuracy and segmentation quality. 
Note that a message size of $4$ is sufficient to achieve comparable performance on both selection and segmentation, after which the performance plateaus. 
We also conduct a similar experiment by varying the key size from $4$ to $1024$, and the same trend can be observed in Figure~\ref{fig:message_key}b. 
Also, we empirically found that a small message (\textit{e.g.,} 8) paired with large key size (\textit{e.g.,} 1024) can achieve amenable performance. 
This is also the sizes of our models for all experiments in Sec.~\ref{sec:results-cases}. 
Importantly, the effectiveness of asymmetric sizes is an interesting finding as it allows us to use a small size for the message that is sent to other agents while using larger sizes for the key used locally by each agent to compute the scores (and hence does not need to be transmitted). These results validate this advantage.


\vspace{-1mm}
\section{Conclusion}
\label{sec:conclusion}
In this paper, we formulated the problem of \textit{collaborative perception}, where agents can combine their local observations with those of other agents in order to improve performance on scene understanding tasks. Inspired by the network communication literature, we propose a handshake communication mechanism for which the network can learn compressed representations. 
Key to our approach is that we decouple the message, key, and value elements to support asymmetric compression, resulting in bandwidth savings. We introduce the \DatasetName~dataset and benchmarking metrics to evaluate our method, and show that our method is able to effectively combine information from neighboring agents to improve accuracy using significantly less bandwidth than centralized approaches.

\vspace{-1mm}
\section{Acknowledgement}
\label{sec:acknowledgement}
This work was supported by ONR grant N00014-18-1-2829.

\clearpage

\printbibliography
\end{document}